# CHIME: Conditional Hallucination and Integrated Multi-scale Enhancement for Time Series Diffusion Model

Yuxuan Chen, Haipeng Xie[*]

*School of Electrical Engineering, Xi'an Jiaotong University, Xi'an, Shaanxi 710049, China*

**Abstract**

The denoising diffusion probabilistic model has become a mainstream generative model, achieving significant success in various computer vision tasks. Recently, there has been initial exploration of applying diffusion models to time series tasks. However, existing studies still face challenges in multi-scale feature alignment and generative capabilities across different entities and long-time scales. In this paper, we propose CHIME, a conditional hallucination and integrated multi-scale enhancement framework for time series diffusion models. By employing multi-scale decomposition and integration, CHIME captures the decomposed features of time series, achieving in-domain distribution alignment between generated and original samples. In addition, we introduce a feature hallucination module in the conditional denoising process, enabling the temporal features transfer across long-time scales. Experimental results on publicly available real-world datasets demonstrate that CHIME achieves state-of-the-art performance and exhibits excellent generative generalization capabilities in few-shot scenarios.

## 1. Introduction

Time series is a common data organization format in the real world (Lim & Zohren, 2021), offering significant research and analytical value across diverse fields such as economics (Hou et al., 2021), engineering (Guan et al., 2021), meteorology (Wu et al., 2023), medicine (Rajpurkar et al., 2022), and transportation (Fang et al., 2022). With the growing challenges of data scarcity and privacy protection, generating time series has emerged as a potential solution. Additionally, time series forecasting plays a crucial role, serving as a key tool for addressing various downstream tasks, including economic dispatch and traffic flow optimization.

Early methods for time series generation were primarily based on GANs, VAEs, and similar approaches, while time series forecasting mostly relied on RNN architectures or Transformer-based models. In recent years, following the significant success of denoising diffusion probabilistic models (DDPM) (Ho et al., 2020), the application of diffusion

---

[*] Corresponding author. Tel.: +86-18991840086; fax: +84-29-8266-5489.
*E-mail address:* haipengxie@xjtu.edu.cn

Click here to enter text.

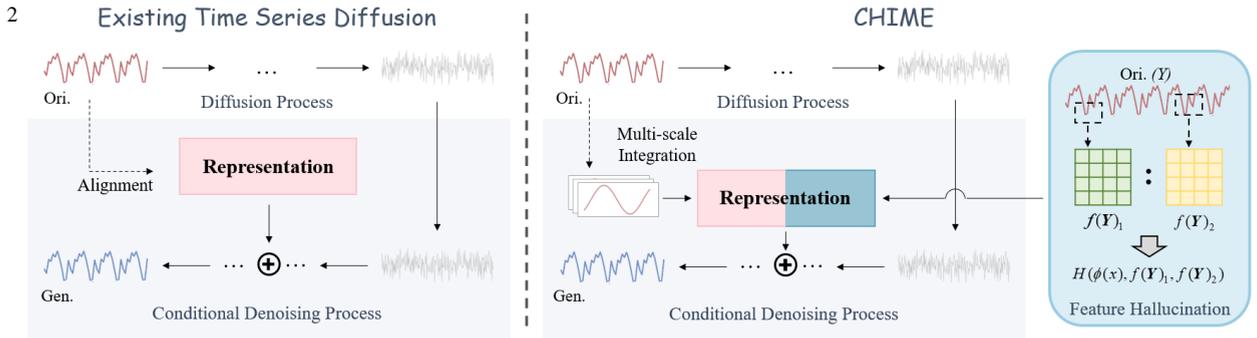

**Fig. 1** CHIME vs. existing time series diffusion models

models to time series tasks has become an emerging research trend. The first attempt was made by Rasul et al. (2021), who integrated the RNN structure into the denoising process to capture the unique semantic meaning of time series. Similarly, CSDI (Tashiro et al., 2021) and SSSD (Lopez Alcaraz & Strodthoff, 2023) introduced non-autoregressive denoising with self-supervised masking and a structured state-space model, respectively, to adapt diffusion models to time series context. However, these regression-based methods struggle to handle long-term time series and fail to model the individual trend and seasonal components effectively. More recently, TimeDiff (Shen & Kwok, 2023) developed a time series diffusion model for long time scales, while Diffusion-TS (Yuan & Qiao) introduced time series decomposition into the denoising diffusion process for the first time. mr-diff (Shen et al., 2024) and MG-TSD (Fan et al.) applied decomposition and generative approaches to extract multi-scale information for conditional denoising. However, their decoupling characteristics lead to error accumulation, limiting the quality of the generated results.

Additionally, real-world time series data presents challenges not addressed by current benchmark datasets, especially in industrial applications. The continuous emergence of new entities introduces richer and more diverse feature variation patterns. Moreover, numerous data-scarce scenarios place higher demands on the transferability of time series models in few-shot settings. While diffusion models show reasonable ability to generalize within context-specific features, they struggle to capture feature transfer over long time scales or to adapt to distribution shifts between different entities.

To address these limitations, this study proposes CHIME, a **c**onditional **h**allucination and **i**ntegrated **m**ulti-scale **e**nhancement framework for time series diffusion models. As shown in Fig. 1, unlike traditional time series diffusion models, the proposed method not only achieves alignment of in-domain feature distributions of generated samples with real samples but also improves the model's generalization ability across a variety of unseen time series domains. This is achieved through two key modules within the CHIME framework. First, we introduce integrated multi-scale enhancement, where multi-scale decomposition is employed to capture the different components within the time series. Through conditioning denoising process for each scale information, we assess various temporal patterns and assign weights to them, which helps alleviate the error accumulation. Second, to obtain accurate representations of long time-scale time series, we propose feature hallucination module in this conditional denoising framework. This module aims to achieve temporal transfer by training a set of category-independent transformation sets, thereby improving the model's overall generative generalization capability.

In summary, the major contributions of this study are as follows:

1) We propose the CHIME framework, a time series diffusion model that combines feature hallucination and multi-scale integration. This approach not only ensures feature alignment between generated and original samples but also demonstrates outstanding few-shot capabilities.

2) We design a multi-scale decomposition and integration framework to capture the characteristics of different time series components, ensuring in-domain distribution alignment. Moreover, we incorporate the Feature Hallucination module into the inference pipeline to facilitate long-range feature adaptation and refinement.

3) Extensive experiments on publicly available time series datasets demonstrate that CHIME outperforms traditional generative models and other diffusion-based methods. Notably, CHIME exhibits superior performance in few-shot scenarios.

## 2. Related Work

### 2.1. Time Series Generation

Traditional generative models, such as Generative Adversarial Networks (GANs), Variational Autoencoders (VAEs), and Normalizing Flows (NFs), have demonstrated reasonable performance across various domains in generative tasks. When applied to time series, a natural extension is to incorporate models with sequential processing capabilities, such as Recurrent Neural Networks (RNNs), into the generative model framework. Esteban et al. (2017) introduced RCGAN, which utilizes RNN structures for both the generator and the discriminator. Mogren (2016) proposed C-RNN-GAN, where random noise is fed into an LSTM to learn hidden representations, which are subsequently passed to the discriminator. Yoon et al. (2019) developed TimeGAN, which employs autoregression to model temporal dependencies in GANs. To generate longer time sequences, Jeha et al. (2022) introduced PSA-GAN, which integrates convolutional layers and self-attention mechanisms into the GAN architecture. TimeVAE offers the added benefit of encoding domain knowledge while providing interpretable generations (Desai et al.). Additionally, Alaa et al. (2021) combined Discrete Fourier Transform with Normalizing Flows to create an explicit likelihood model.

### 2.2. Time Series Forecasting

Traditional time series forecasting methods utilize mathematical and statistical theories, such as ARIMA, Seasonal ARIMA (SARIMA), and Exponential Smoothing (ETS) models. While these methods offer advantages such as theoretical simplicity and compact model sizes, they often struggle to capture the complex nonlinear patterns inherent in time series data.

For deep models, RNNs architectures such as LSTM are effective at capturing sequential and contextual dependencies in time series data. In recent years, Transformer-based (Vaswani, 2017) models have excelled in time-series tasks. Zhou et al. (2021) introduced Informer, which utilizes sparse attention to enhance the inference speed of Transformers. Wu et al. (2021) developed Autoformer, which incorporates auto-correlation to enhance its progressive decomposition ability. Fedformer (Zhou et al., 2022) adopts frequency-domain mapping to better learn the decomposition representation of time series. (Nie et al.) proposed PatchTST, which divides time series into patches to learn local semantics. More recently, Zeng et al. (2023) introduced DLinear, which leverages a single linear layer combined with seasonal trend decomposition for training, achieving excellent prediction performance.

### 2.3. Time Series Diffusion Models

Recently, diffusion-based generative models have been applied to time series tasks, improving sample quality and offering greater robustness by gradually denoising data in a controlled manner. TimeGrad (Rasul et al., 2021) was the first to apply diffusion to time series tasks, using an RNN structure to guide conditional denoising. CSDI (Tashiro et al., 2021) adopts a self-supervised, non-autoregressive strategy for generation, while SSSD (Lopez Alcaraz & Strodthoff, 2023) replaces the Transformer architecture with a structured state-space model. TimeDiff (Shen & Kwok, 2023) models long-term dependencies by introducing future mix-up. More recently, Diffusion-TS (Yuan & Qiao) introduces trend-seasonal decomposition of time series into diffusion models. Shen et al. (2024) proposed mr-diff, which applies multi-scale decomposition to guide the generation of high-quality samples. However, this approach completely decouples different scales, exacerbating error accumulation. (Fan et al.) introduced MG-TSD, constructing a multi-granularity guidance diffusion loss. However, this method is designed for fixed observation prediction tasks, limiting its generalization ability and hindering the modeling of long-term scale distributions. Additionally, the aforementioned works focus primarily on modeling more accurate features for the conditional guidance process, ensuring that the generated samples align with the distribution of historical samples. Yet they overlook the potential intra-class feature distribution shifts across different subjects or over extended time scales.

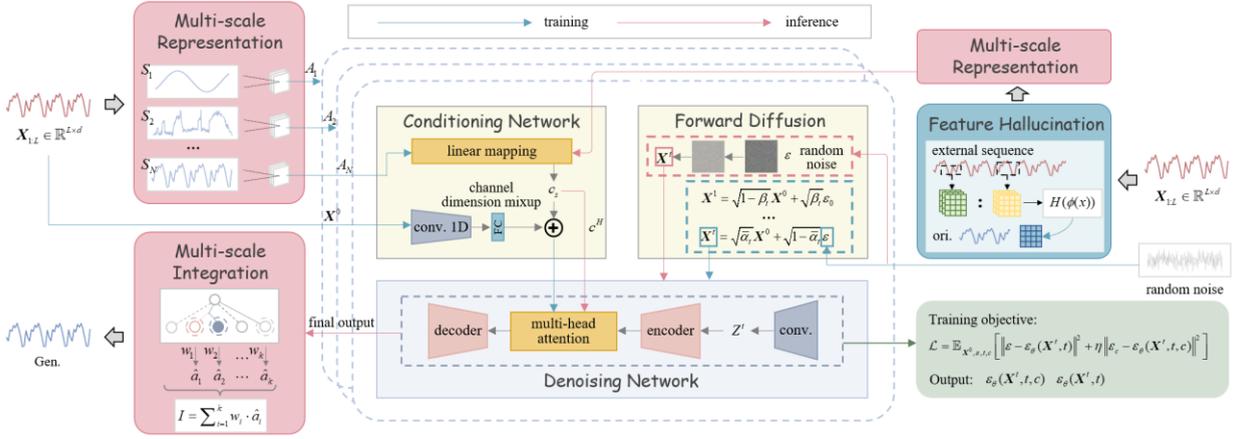

**Fig. 2** The overall framework of our proposed model

## 3. Methodology

### 3.1. Problem Statement

**Time Series Generation.** Let $\boldsymbol{X}_{Ori} = \boldsymbol{X}_{1:L} = \{x_1, ..., x_L\} \in \mathbb{R}^{L \times d}$ denote the observable time series, where $L$ represents the number of time steps, and $d$ denotes the number of channels in a multivariate time series. Our goal is to pre-train a diffusion-based model guided by conditional information $p(\cdot \mid c)$, ultimately producing a generated time series $\boldsymbol{X}_{Gen}$ that is most similar to the original time series $\boldsymbol{X}_{Ori}$.

**Time Series Forecasting.** For the long-term forecasting task, given a time series $\boldsymbol{X}_{Ori} = \boldsymbol{X}_{1:L} = \{x_1, ..., x_L\} \in \mathbb{R}^{L \times d}$ with a lookback window of length $L$, our objective is to obtain the predicted values for the next $h$ steps $\boldsymbol{X}_{Pre} = \boldsymbol{X}_{L+1:L+h} = \{x_{L+1}, ..., x_{L+h}\} \in \mathbb{R}^{h \times d}$. We approach the time series forecasting problem from a generative perspective, learning a density $p_\theta(\boldsymbol{X}_{Pre} \mid \boldsymbol{X}_{Ori})$ in our conditional diffusion framework that best approximates the real distribution $p(\boldsymbol{X}_{Pre} \mid \boldsymbol{X}_{Ori})$.

### 3.2. CHIME Framework

This paper proposes CHIME, a diffusion-based framework that integrates multi-scale enhancement and feature hallucination, as illustrated in Fig. 2. To address the limitation of existing time series diffusion models that fail to effectively incorporate multi-scale information, CHIME achieves detailed modeling of in-domain feature distributions through multi-scale decomposition and integration. Simultaneously, feature hallucination is employed in inference phase to facilitate cross-subject feature transfer across long time scales, thereby enhancing the model's generalization capability and few-shot learning ability.

Furthermore, CHIME is based on a conditional denoising framework. We design a novel training paradigm for conditional denoising, which distinguishes itself from existing approaches (Liu et al., 2024; Zhang et al., 2023) that simply adopt a global attention architecture, causing the model to focus continuously on the reference during the denoising process. In contrast, our approach applies a pipeline consisting of three modules for each scale information. First, the scale information and the feature embedding of the original sequence pass through the conditioning network to generate conditional information that guides the denoising network. Meanwhile, after passing through the forward diffusion module the noised sequence is input into the denoising network. After going through a convolutional layer and encoder, it is combined with the conditional information. Then, the decoder outputs $\varepsilon_\theta$. The output for each scale's pipeline is further processed through the multi-scale integration module to obtain the final generated sequence. Experimental results demonstrate that our designed conditional diffusion framework achieves excellent performance.

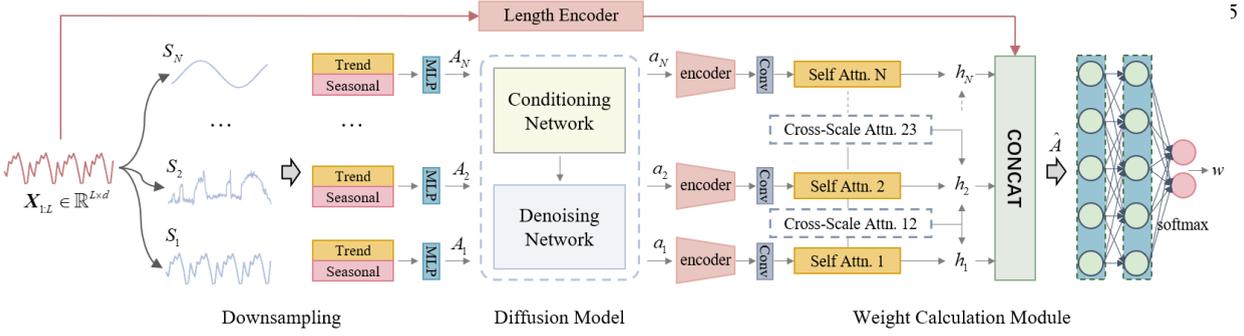

**Fig. 3** Multi-scale integration framework

### 3.3. Integrated Multi-scale Enhancement for Time Series

Modeling the different components of a time series is crucial for obtaining more refined feature representations. Although recent work (Fan et al.; Shen et al., 2024; Yuan & Qiao) has explored the decomposition of time series in diffusion processes, it overlooks the interaction between components at different scales and fails to account for the drift in multi-scale weight distributions. In reality, most real-world time series exhibit interacting multi-scale components. For instance, electricity load time series demonstrate fine-grained daily variations at scales sampled at 15-minute or 1-hour intervals, while trends sampled on a monthly basis influence the baseline of these fine-grained variations. Moreover, monthly data is affected by the overarching annual cycle.

This paper proposes a multi-scale decomposition and integration paradigm, as shown in Fig .3. First, rather than relying on traditional time series trend-seasonal decomposition, we learn a set of different downsampling rates to decompose the time series into representations across multiple scales. Next, for each downsampled feature sequence, we treat it as the input to the time series diffusion model for training. Finally, a multi-scale weight calculation module is introduced to determine the contribution of each component, facilitating the integration of multi-scale information.

Given a history time series $X_{1:L} = \{x_1,...,x_L\} \in \mathbb{R}^{L \times d}$ and a set of downsampling rate $\{s_1, s_2,...,s_N\}$, the downsampling and encoder process can be expressed as:

$$Downsampling : \{S_i\} \quad S_i(X_{1:L}) = \left\{ \frac{1}{s_i} \sum_{j=1}^{m} X_{(j-1) \cdot s_i : j \cdot s_i} \right\}, \ m = \lfloor L / s_i \rfloor \quad (1)$$

$$\{A_i\} = \{MLP(Trend(S_i)) + MLP(Seasonal(S_i))\}, \quad i = 1, 2,..., N \quad (2)$$

where, $S_i$ represents the data obtained after downsampling with a sampling rate of $s_i$, and trend-seasonal decomposition is performed on multi-scale information at different granularities. After passing through an MLP module, we obtain the preliminary representation of the multi-scale patterns.

After that, the divided multi-scale information is used as conditional information and input into the conditioning network of the diffusion model, which then outputs the generated multi-scale series.

Finally, we input the generated multi-scale information into the weight calculation module. Through weighted integration in temporal dimension, the results of the generative model are made to more closely approximate the original features:

$$Weight : \quad w = \text{softmax}((W\hat{A} + b) / \tau) \quad (3)$$

$$Integration : \quad I = \sum_{i=1}^{k} w_i \cdot \hat{a}_i \quad (4)$$

where $\tau$ represents learnable temperature parameter. The integrated mixed-scale information $I$ is composed of weighted parts, where $\hat{a}_i$ represents the output scale information of time series diffusion model.

In the overall CHIME framework, the multi-scale integration module runs throughout the entire training process. The weight calculation module for the multi-scale information is trained globally alongside the diffusion model. Additionally, the output $\{A\}$ from the multi-scale decomposition and feature extraction is primarily used to construct the input for the conditioning network. Detailed model specifications can be found in Section 3.5.

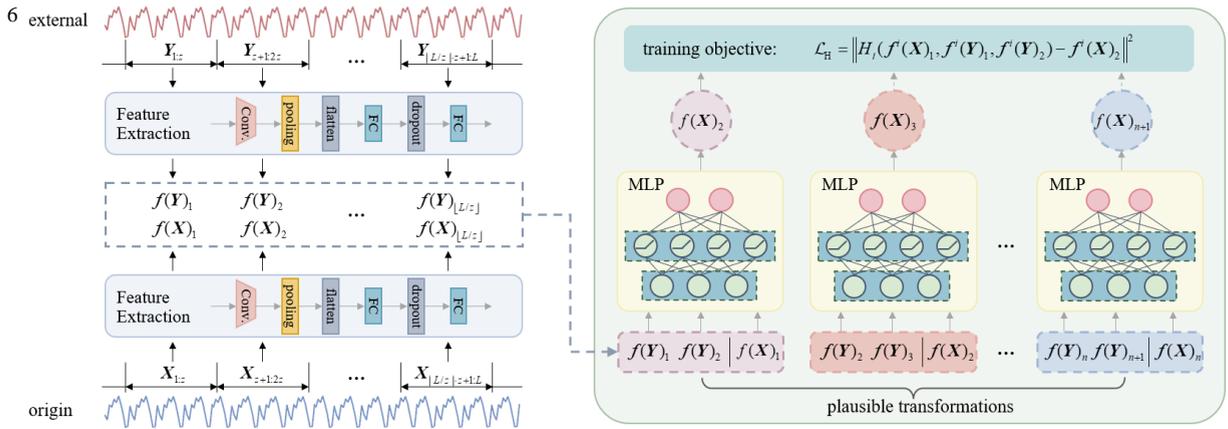

**Fig. 4** Feature Hallucination module

### 3.4. Feature Hallucination for Enhanced Temporal Transfer

Previous work has often overlooked the potential distribution shifts caused by cross-subject generation and few-shot settings encountered in practical applications. For better generalization ability of diffusion and improved generation quality, we have incorporated a feature hallucination module within the CHIME framework. General feature extraction is performed using an additional time series set from external sources, transferring the high-dimensional semantic representations of the external sequences to the target sequence.

Hariharan and Girshick (2017) highlighted a critical issue in visual recognition tasks within the computer vision field. When intra-class variations are not learned during the training phase, classifiers are likely to draw incorrect conclusions. For instance, if the training phase only includes a specific type of bird in flight, the classifier may lose the ability to correctly identify the same bird species perching on a tree. Therefore, they designed a feature hallucination approach to transfer the resting state of other bird species to this particular species, thereby obtaining more complete training samples. Inspired by this idea, we propose a key hypothesis: Is the phenomenon of intra-class feature distribution shifts naturally aligned with the changes in time-series data across different subjects and long-time scales? Since pre-trained diffusion models rely on conditional information to guide the sampling process, we believe it is essential to introduce feature hallucination to refine the guiding information during the denoising phase.

In the field of time series, we consider a realistic scenario where a limited amount of data is available, creating a few-shot setting. This data often exhibits changes over long-time scales, such as stock prices and electric load data. For example, they exhibit completely different baselines across seasons within a year, but share similar local fluctuation patterns throughout the year. However, the scarcity of available data makes it difficult for models to capture such long-term variations, limiting its ability to learn only local feature changes. In this condition, directly applying a pre-trained unified diffusion model typically fails to sample with feature distributions accurately reflect the actual data. Retraining a diffusion model on this limited data risks losing feature variations on even longer time scales. To address this, our approach leverages existing time series data from the same category such as data from other stocks or load sources for feature hallucination. The key to this process is identifying a set of category-independent transformations that can be applied to the conditional information, leading to a more accurate guidance of the denoising process.

To achieve this, we designed an MLP-based structure, denoted as $H(\cdot)$. We hypothesize that there exists a feature distribution hallucination between any two non-overlapping segments within the same category of time series, such as $X_{1:l_1}$, $X_{l_2:l_3}$. For a new time series $X^{\text{new}}$, if $H(\cdot)$ can produce an output $X^H$ from the input "$X_{1:l_1}$, $X_{l_2:l_3}$, $X^{\text{new}}$" such that $X^{\text{new}} : X^H$ is as close as possible to $X_{1:l_1} : X_{l_2:l_3}$, then we consider $H(\cdot)$ have successfully hallucinated the features of the time series segment $X^{\text{new}}$.

The proposed Feature Hallucination module is shown in Fig. 4. First, we need to obtain a sufficient and suitable set of transformations, such as $X_{1:l_1} : X_{l_2:l_3}$. We define a set of segmentation resolutions with varying granularities $\{z_1, z_2, ..., z_n\}$, to determine the division of regions, which allows us to generate a series of sequences $Z_n(x)$. Taking $Z_i$ as an example:

$$Z_i(\boldsymbol{X}_{1:L}) = \left\{ \boldsymbol{X}_{1:z_i}, \boldsymbol{X}_{z_i+1:2z_i}, ..., \boldsymbol{X}_{\lfloor L/z_i \rfloor z_i+1:L} \right\} \tag{5}$$

Each time series segment is first encoded into a feature representation $\{f^i(\boldsymbol{X})_j\}$, $j = 1,...,\lfloor L/z_i \rfloor$ using an encoder composed of convolutional and fully connected layers. We repeat the above process on external time series $\boldsymbol{Y}_{1:L}$ to obtain a corresponding set of $\{f^i(\boldsymbol{Y})_j\}$, $j = 1,...,\lfloor L/z_i \rfloor$. Next, we select one pair of $(f^i(\boldsymbol{X})_1, f^i(\boldsymbol{X})_2)$ and one pair of $(f^i(\boldsymbol{Y})_1, f^i(\boldsymbol{Y})_2)$ such that the cosine similarity between $f^i(\boldsymbol{X})_2 - f^i(\boldsymbol{X})_1$ and $f^i(\boldsymbol{Y})_2 - f^i(\boldsymbol{Y})_1$ is minimized. This set of $(f^i(\boldsymbol{X})_1, f^i(\boldsymbol{X})_2, f^i(\boldsymbol{Y})_1, f^i(\boldsymbol{Y})_2)$ is then added to the training set of $H_i(\cdot)$. Taking $(f^i(\boldsymbol{X})_1, f^i(\boldsymbol{Y})_1, f^i(\boldsymbol{Y})_2)$ as input, $H_i(\cdot)$ outputs $\hat{f}^i(\boldsymbol{X})_2$. The loss function is:

$$\mathcal{L}_H = \left\| H_i(f^i(\boldsymbol{X})_1, f^i(\boldsymbol{Y})_1, f^i(\boldsymbol{Y})_2) - f^i(\boldsymbol{X})_2 \right\|^2 \tag{6}$$

After training, we obtain a set of $\{H_i\}$, which exhibits feature hallucination capabilities at different scales. In practical applications, we adopt two approaches: 1) For actual scales included in the predefined set $\{z_1, z_2,..., z_n\}$ (e.g., weekly or monthly data in load data), we directly use the corresponding $H_i$ for feature hallucination; 2) If the actual scale is not in $\{z_1, z_2,..., z_n\}$, we apply a weighted processing using the nearest $H$. After feature hallucination process, the conditioning network will output $c^H$, which serves as the guidance in denoising process, leading to more accurate sampling capabilities.

### 3.5. Conditional Diffusion for Time Series Generation

The DDPM model, proposed by Ho et al. (2020), represents a ground-breaking advancement in generative models within the computer vision field. It works by progressively adding Gaussian noise to the data and then training a reverse denoising process. In this paper, we explore the generation of time series from a diffusion perspective. Given an original time series $\boldsymbol{X}_{Ori} = \boldsymbol{X}_{1:L} = \{x_1,...,x_L\} \in \mathbb{R}^{L \times d}$ and a generated time series $\boldsymbol{X}_{Gen}$, our goal is to learn a denoising network that minimizes the distance between the distribution $p_\theta(\boldsymbol{X}_{Gen} \mid \boldsymbol{X}_{Ori})$ and the true conditional distribution $p(\boldsymbol{X}_{Gen} \mid \boldsymbol{X}_{Ori})$ as much as possible.

According to the principles of DDPM, the forward diffusion process of the time series is viewed as a noise addition process with a Markov structure, denoted as: $\boldsymbol{X}_{Ori}^0,..., \boldsymbol{X}_{Ori}^T$. The forward process from state $\boldsymbol{X}^{t-1}$ to $\boldsymbol{X}^t$, denoted as $q$, can be expressed as:

$$q(\boldsymbol{X}^t \mid \boldsymbol{X}^{t-1}) := N(\boldsymbol{X}^t; \sqrt{1-\beta_t} \boldsymbol{X}^{t-1}, \beta_t \boldsymbol{I}) \tag{7}$$

$$q(\boldsymbol{X}^t \mid \boldsymbol{X}^0) := N(\boldsymbol{X}^t; \sqrt{\bar{\alpha}_t} \boldsymbol{X}^0, (1-\bar{\alpha}_t)\boldsymbol{I}) \tag{8}$$

where $\beta_t \in [0,1]$ represents the variance of diffusion step $t$. $\bar{\alpha}_t = \prod_{s=1}^t \alpha_s$, $\alpha_t = 1 - \beta_t$. Thus, $\boldsymbol{X}^t$ can be expressed as:

$$\boldsymbol{X}^t = \sqrt{\bar{\alpha}_t} \boldsymbol{X}^0 + \sqrt{1-\bar{\alpha}_t} \varepsilon, \varepsilon \in N(0, \boldsymbol{I}) \tag{9}$$

The reverse denoising process can still be viewed as a Markov structure, where the state transition from $\boldsymbol{X}^t$ to $\boldsymbol{X}^{t-1}$ can be expressed as:

$$p_\theta(\boldsymbol{X}^{t-1} \mid \boldsymbol{X}^t) = N(\boldsymbol{X}^{t-1}; \mu_\theta(\boldsymbol{X}^t, t), \Sigma_\theta(\boldsymbol{X}^t, t)) \tag{10}$$

where the mean $\mu_\theta$ is obtained through a neural network, $\Sigma_\theta$ is fixed as $\sigma_t^2 \boldsymbol{I}$.

Different from Diffusion-TS, we aim to reconstruct the noise rather than directly predict data. Specifically, based on the posterior distribution, a denoising network $\varepsilon_\theta$ is used to estimate noise with the mean being expressed as:

$$\mu_\theta(\boldsymbol{X}^t, t) = \sqrt{\alpha_t} \boldsymbol{X}^t - \frac{1-\alpha_t}{\sqrt{1-\bar{\alpha}_t} \sqrt{\alpha_t}} \varepsilon_\theta(\boldsymbol{X}^t, t) \tag{11}$$

To train $\varepsilon_\theta$, the loss function is set as:

$$\mathcal{L}(\varepsilon) = \sum_{t=1}^T \mathbb{E}_{\boldsymbol{X}^0, \varepsilon, t} \left\| \varepsilon - \varepsilon_\theta(\boldsymbol{X}^t, t) \right\|^2 \tag{12}$$

In the conditional denoising process, the conditional posterior distribution can be expressed as:

$$p(\boldsymbol{X}^{0:T} \mid c^H) = p(\boldsymbol{X}^T \mid c^H) \prod_{t=1}^T p_\theta(\boldsymbol{X}^{t-1} \mid \boldsymbol{X}^t, c^H) \tag{13}$$

where $p(\boldsymbol{X}^T \mid c^H)$ can be approximated as $N(\boldsymbol{X}^T \mid c^H, \boldsymbol{I})$, and $p_\theta(\boldsymbol{X}^{t-1} \mid \boldsymbol{X}^t, c^H)$ is the conditional transition process with learnable parameters $\theta$. The denoising process at step $t$ can be expressed as:

$$p_\theta(\boldsymbol{X}^{t-1} \mid \boldsymbol{X}^t, c^H) = N(\boldsymbol{X}^{t-1}; \mu_\theta(\boldsymbol{X}^t, c^H, t), \Sigma_\theta(\boldsymbol{X}^t, c^H, t)) \tag{14}$$

where $\Sigma_\theta(\boldsymbol{X}^t, c^H, t)$ can be treated as a fixed value $\sigma_t^2 \boldsymbol{I}$, $\mu_\theta(\boldsymbol{X}^t, c^H, t)$ is the network we design to guide the conditional denoising process and $\theta$ represents the learnable parameters.

To maximize the guiding effect of the conditional information, similar to the work of Zhang et al. (2023), we adopt a multi-head cross-attention structure to the conditional information, as illustrated in Fig. x. Specifically, we consider a trade-off strategy in the denoising network:

$$\nabla_{\boldsymbol{X}^{T-1}} \log p(\boldsymbol{X}^{T-1} \mid \boldsymbol{X}^T, c) = \nabla_{\boldsymbol{X}^{T-1}} \log p(\boldsymbol{X}^{T-1} \mid \boldsymbol{X}^T) + \nabla_{\boldsymbol{X}^{T-1}} \log p(c \mid \boldsymbol{X}^{T-1}) \tag{15}$$

$$\mathcal{L} = \mathbb{E}_{\boldsymbol{X}^0, \varepsilon, c} \left[ \left\| \varepsilon - \varepsilon_\theta(\boldsymbol{X}^t, t) \right\|^2 + \eta \left\| \varepsilon_c - \varepsilon_\theta(\boldsymbol{X}^t, t, c) \right\|^2 \right] \tag{16}$$

## 4. Experiments

### 4.1. Datasets

This paper conducts extensive experiments on 8 publicly available real-world datasets and 2 simulated datasets. The datasets ETTh1, Stocks, Energy, fMRI, MuJoCo, and Sines are used for generative task, while ETTh1, ETTm1, Weather, Traffic, and Electricity are used for predictive task. The ETTh and ETTm datasets record transformer sensor data collected every hour and every 15 minutes over two years, respectively. The Stocks dataset contains daily stock data for Google from 2004 to 2019. The Energy dataset includes forecast data from the UCI appliance dataset. Following the previous study (Yuan & Qiao), the fMRI dataset contains blood oxygen level-dependent (BOLD) data with 50 features. The Sines dataset is a synthetic sine wave simulation, while the MuJoCo dataset is a multivariate physical simulation dataset. The Traffic dataset records hourly traffic flow on San Francisco freeways, and the Weather dataset includes meteorological data collected every 10 minutes from 2020 to 2021, with 21 features. The Electricity dataset consists of hourly load data for 321 users over two years.

### 4.2. Implementation Details

In this study, we use linear noise scheduling, ranging from $\beta_1 = 10^{-4}$ to $\beta_T = 5 \times 10^{-2}$. For the ETTh, Sines, and Stocks datasets, the number of timesteps is set to 500, while for the remaining datasets, it is set to 1000. The Adam optimizer is employed with a base learning rate of $10^{-4}$, and batch size is set to 128. During the conditional denoising process, the number of attention heads in the cross-attention structure is set to 4. All experiments are conducted on a server equipped with 4 Intel Xeon Gold 6254 processors and 2 NVIDIA A800 GPUs. It is worth noting that the feature hallucination process is trained exclusively on the training set, with no information from the test set being accessed beforehand.

### 4.3. Evaluation Metrics

This paper designs specific evaluation metrics for time series generation and prediction tasks. For the generation task, we use the Context-Fréchet Inception Distance (Context-FID) to measure the difference between the generated and original time series representations (Jeha et al., 2022). We also compute the correlation matrix to assess the correlation between the generated and real sequences(Liao et al., 2020).

$$\frac{1}{10} \sum_{i,j} \frac{\text{cov}_{i,j}^{\text{Ori}}}{\sqrt{\text{cov}_{i,i}^{\text{Ori}} \cdot \text{cov}_{j,j}^{\text{Ori}}}} - \frac{\text{cov}_{i,j}^{\text{Gen}}}{\sqrt{\text{cov}_{i,i}^{\text{Gen}} \cdot \text{cov}_{j,j}^{\text{Gen}}}} \tag{17}$$

Additionally, following the experimental setup of (Yuan & Qiao), we use a network based on the GRU architecture to further evaluate the practical utility of the generated data. The discriminative and predictive capabilities of the generated data are assessed through supervised classification and prediction tasks.

For the time series prediction task, we adopt two commonly used error metrics: Mean Absolute Error (MAE) and Mean Squared Error (MSE).

*4.4. Time Series Generation Results*

In this section, we evaluate the time series generation capabilities of CHIME on six datasets (Energy, ETTh, fMRI, MuJoCo, Sines, Stocks), with the results presented in Table 1. The best results are highlighted in bold.

**Table 1** Time Series Generation Results (24-length). Results for all baseline models are from (Yuan & Qiao)

| | | CHIME | Diffusion-TS | TimeGAN | TimeVAE | Diffwave | DiffTime | Cot-GAN |
|---|---|---|---|---|---|---|---|---|
| Sines | Context-FID | **0.005±.000** | 0.006±.000 | 0.101±.014 | 0.307±.060 | 0.014±.002 | 0.006±.001 | 1.337±.068 |
| | Correlation | **0.014±.005** | 0.015±.004 | 0.045±.010 | 0.131±.010 | 0.022±.005 | 0.017±.004 | 0.049±.010 |
| | Discriminative | **0.006±.005** | 0.006±.007 | 0.011±.008 | 0.041±.044 | 0.017±.008 | 0.013±.006 | 0.254±.137 |
| | Predictive | **0.091±.000** | 0.093±.000 | 0.093±.019 | 0.093±.000 | 0.093±.000 | 0.093±.000 | 0.100±.000 |
| Stocks | Context-FID | 0.149±.034 | 0.147±.025 | **0.103±.013** | 0.215±.035 | 0.232±.032 | 0.236±.074 | 0.408±.086 |
| | Correlation | **0.003±.001** | 0.004±.001 | 0.063±.005 | 0.095±.008 | 0.030±.020 | 0.006±.002 | 0.087±.004 |
| | Discriminative | 0.075±.019 | **0.067±.015** | 0.102±.021 | 0.145±.120 | 0.232±.061 | 0.097±.016 | 0.230±.016 |
| | Predictive | **0.034±.000** | 0.036±.000 | 0.038±.001 | 0.039±.000 | 0.047±.000 | 0.038±.001 | 0.047±.001 |
| ETTh | Context-FID | **0.105±.007** | 0.116±.010 | 0.300±.013 | 0.805±.186 | 0.873±.061 | 0.299±.044 | 0.980±.071 |
| | Correlation | **0.044±.009** | 0.049±.008 | 0.210±.006 | 0.111±.020 | 0.175±.006 | 0.067±.005 | 0.249±.009 |
| | Discriminative | **0.057±.010** | 0.061±.009 | 0.114±.055 | 0.209±.058 | 0.190±.008 | 0.100±.007 | 0.325±.099 |
| | Predictive | **0.118±.002** | 0.119±.002 | 0.124±.001 | 0.126±.004 | 0.130±.001 | 0.121±.004 | 0.129±.000 |
| MuJoCo | Context-FID | **0.012±.001** | 0.013±.001 | 0.563±.052 | 0.251±.015 | 0.393±.041 | 0.188±.028 | 1.094±.079 |
| | Correlation | **0.165±.036** | 0.193±.027 | 0.886±.039 | 0.388±.041 | 0.579±.018 | 0.218±.031 | 1.042±.007 |
| | Discriminative | **0.007±.002** | 0.008±.002 | 0.238±.068 | 0.230±.102 | 0.203±.096 | 0.154±.045 | 0.426±.022 |
| | Predictive | **0.006±.000** | 0.007±.000 | 0.025±.003 | 0.012±.002 | 0.013±.000 | 0.010±.001 | 0.068±.009 |
| Energy | Context-FID | **0.081±.026** | 0.089±.024 | 0.767±.103 | 1.631±.142 | 1.031±.131 | 0.279±.045 | 1.039±.028 |
| | Correlation | **0.697±.136** | 0.856±.147 | 4.010±.104 | 1.688±.226 | 5.001±.154 | 1.158±.095 | 3.164±.061 |
| | Discriminative | **0.109±.003** | 0.122±.003 | 0.236±.012 | 0.499±.000 | 0.493±.004 | 0.445±.004 | 0.498±.002 |
| | Predictive | **0.250±.000** | 0.250±.000 | 0.273±.004 | 0.292±.000 | 0.251±.000 | 0.252±.000 | 0.259±.000 |
| fMRI | Context-FID | **0.102±.005** | 0.105±.006 | 1.292±.218 | 14.449±.969 | 0.244±.018 | 0.340±.015 | 7.813±.550 |
| | Correlation | **1.402±.042** | 1.411±.042 | 23.502±.039 | 17.296±.526 | 3.927±.049 | 1.501±.048 | 26.824±.449 |
| | Discriminative | **0.158±.020** | 0.167±.023 | 0.484±.042 | 0.476±.044 | 0.402±.029 | 0.245±.051 | 0.492±.018 |
| | Predictive | **0.088±.000** | 0.099±.000 | 0.126±.002 | 0.113±.003 | 0.101±.000 | 0.100±.000 | 0.185±.003 |
| Average | Context-FID | **0.075±.013** | 0.079±.011 | 0.521±.069 | 2.943±.235 | 0.465±.048 | 0.225±.035 | 2.112±.147 |
| | Correlation | **0.387±.038** | 0.421±.038 | 4.786±.034 | 3.285±3.469 | 1.622±.042 | 0.495±.031 | 5.236±.090 |
| | Discriminative | **0.060±.017** | 0.072±.010 | 0.198±.034 | 0.267±.061 | 0.256±.034 | 0.176±.022 | 0.371±.049 |
| | Predictive | **0.097±.000** | 0.101±.000 | 0.113±.005 | 0.113±.002 | 0.113±.002 | 0.102±.001 | 0.131±.002 |

From the table, it is clear that CHIME outperforms all other baseline models in terms of metric scores across nearly all datasets, achieving state-of-the-art performance in time series generation. Furthermore, the comparative results demonstrate that diffusion-based models consistently outperform traditional GAN-based generation methods in time series tasks.

Figure 5 provides two types of visualizations for the generated samples from CHIME. The upper section shows the t-SNE dimensionality reduction results for both the original and generated samples, while the lower section illustrates the distribution of the original and generated time series. It is evident that the time series generated by CHIME exhibit similar shapes in high-dimensional features as the original time series and effectively cover the entire range of the original samples. This validates the effectiveness of the multi-scale integration modeling and feature hallucination techniques proposed in this paper.

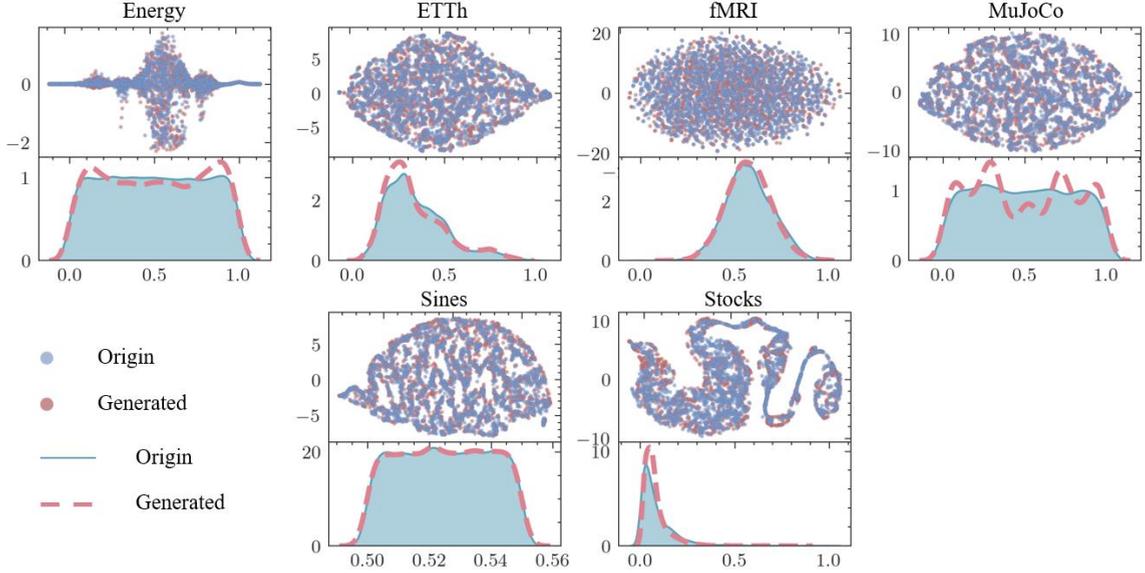

**Fig. 5** Visualization of CHIME time series generation results (t-SNE & kde)

*4.5. Time Series Forecasting Results*

In this section, we evaluate the time series forecasting capability of CHIME on five datasets: ETTh1, ETTm1, Weather, Electricity, and Traffic. Table 2 presents the univariate time series forecasting results across multiple time scales, comparing CHIME with both generative time series models and other deep learning-based models.

**Table 2** Univariate time series forecasting results. Results are averaged over prediction lengths $\{96, 192, 336, 720, 1440\}$. Results for all baselines are from (Shen et al., 2024); best results are in bold.

| | ETTh1 | | ETTm1 | | Weather | | Electricity | | Traffic | | Avg rank |
|---|---|---|---|---|---|---|---|---|---|---|---|
| | MSE | MAE | MSE | MAE | MSE | MAE | MSE | MAE | MSE | MAE | |
| CHIME | **0.065**(1) | 0.199(2) | **0.037**(1) | **0.145**(1) | **0.002**(1) | 0.036(4) | **0.230**(1) | **0.327**(1) | **0.119**(1) | **0.192**(1) | **1.4** |
| mr-diff | 0.066(2) | **0.196**(1) | 0.039(2) | 0.149(2) | **0.002**(1) | **0.032**(1) | 0.234(3) | 0.332(2) | **0.119**(1) | 0.197(2) | 1.7 |
| TimeDiff | 0.066(2) | 0.202(3) | 0.040(3) | 0.154(4) | **0.002**(1) | 0.035(2) | 0.232(2) | 0.341(3) | 0.121(3) | 0.207(3) | 2.6 |
| TimeGrad | 0.078(6) | 0.212(4) | 0.048(7) | 0.167(7) | **0.002**(1) | 0.036(4) | 0.920(14) | 0.898(14) | 1.223(14) | 0.894(14) | 3.8 |
| CSDI | 0.083(11) | 0.221(8) | 0.050(10) | 0.170(9) | **0.002**(1) | 0.037(6) | 0.520(10) | 0.540(11) | 0.393(12) | 0.468(12) | 9.0 |
| SSSD | 0.097(13) | 0.250(13) | 0.049(8) | 0.169(8) | 0.004(9) | 0.041(9) | 0.370(7) | 0.403(6) | 0.151(5) | 0.226(4) | 8.2 |
| D³VAE | 0.078(6) | 0.221(8) | 0.044(6) | 0.160(6) | 0.003(7) | 0.039(7) | 0.535(11) | 0.539(10) | 0.151(5) | 0.237(6) | 7.2 |
| CPF | 0.080(9) | 0.221(8) | 0.041(4) | 0.153(3) | **0.002**(1) | 0.035(2) | 0.609(12) | 0.592(12) | 0.279(11) | 0.385(11) | 7.3 |
| PSA-GAN | 0.084(12) | 0.225(12) | 0.051(11) | 0.174(11) | 0.035(13) | 0.139(13) | 0.273(5) | 0.373(5) | 0.250(10) | 0.355(10) | 10.2 |
| Dlinear | 0.078(6) | 0.222(11) | 0.041(4) | 0.155(5) | 0.168(14) | 0.310(14) | 0.244(4) | 0.348(4) | 0.139(4) | 0.236(5) | 7.1 |
| PatchTST | 0.106(14) | 0.260(14) | 0.052(13) | 0.174(11) | 0.026(12) | 0.098(12) | 0.450(9) | 0.478(9) | 0.177(8) | 0.269(7) | 10.9 |
| Fedformer | 0.076(4) | 0.212(4) | 0.065(14) | 0.195(14) | 0.005(10) | 0.057(10) | 0.376(8) | 0.453(8) | 0.173(7) | 0.278(8) | 8.7 |
| Autoformer | 0.081(10) | 0.220(7) | 0.051(11) | 0.174(11) | 0.003(7) | 0.040(8) | 0.659(13) | 0.623(13) | 0.473(13) | 0.495(13) | 10.6 |
| Informer | 0.076(4) | 0.212(4) | 0.049(8) | 0.172(10) | 0.007(11) | 0.069(11) | 0.363(6) | 0.433(7) | 0.213(9) | 0.308(9) | 7.9 |

To ensure fairness in evaluation, CHIME was provided with the same length of observed continuous time points as conditions for all time scales, consistent with other deep learning-based models, and was then tasked with forecasting the specified future steps. As shown in the table, CHIME achieves the best prediction performance on all five datasets, outperforming state-of-the-art diffusion-based and deep learning-based models.

**Table 3** Multivariate time series forecasting results. Results are averaged over prediction lengths $\{96, 192, 336, 720, 1440\}$. Results of all baselines are from (Shen et al., 2024); best results are in bold.

| | ETTh1 | | ETTm1 | | Weather | | Electricity | | Traffic | | Avg rank |
|---|---|---|---|---|---|---|---|---|---|---|---|
| | MSE | MAE | MSE | MAE | MSE | MAE | MSE | MAE | MSE | MAE | |
| CHIME | **0.403**(1) | **0.415**(1) | **0.328**(1) | **0.368**(1) | **0.294**(1) | 0.334(3) | **0.147**(1) | **0.245**(1) | 0.412(2) | 0.279(2) | **1.4** |
| mr-diff | 0.411(3) | 0.422(2) | 0.340(3) | 0.373(3) | 0.296(2) | 0.324(2) | 0.155(2) | 0.253(2) | 0.474(3) | 0.320(3) | 2.5 |
| TimeDiff | 0.407(2) | 0.430(3) | 0.336(2) | 0.372(2) | 0.311(3) | **0.312**(1) | 0.193(3) | 0.305(3) | 0.564(4) | 0.384(4) | 2.7 |
| TimeGrad | 0.993(14) | 0.719(14) | 0.874(14) | 0.605(14) | 0.392(10) | 0.381(9) | 0.736(12) | 0.630(12) | 1.745(13) | 0.849(13) | 12.5 |
| CSDI | 0.497(5) | 0.438(4) | 0.529(10) | 0.442(11) | 0.356(6) | 0.374(7) | / | / | / | / | 7.2 |
| SSSD | 0.726(11) | 0.561(11) | 0.464(8) | 0.406(8) | 0.349(5) | 0.350(5) | 0.255(8) | 0.363(7) | 0.642(6) | 0.398(7) | 7.6 |
| D³VAE | 0.504(6) | 0.502(9) | 0.362(5) | 0.391(5) | 0.375(8) | 0.380(8) | 0.286(9) | 0.372(8) | 0.928(10) | 0.483(10) | 7.8 |
| CPF | 0.730(12) | 0.597(13) | 0.482(9) | 0.391(6) | 1.390(14) | 0.781(14) | 0.793(13) | 0.643(13) | 1.625(12) | 0.714(12) | 11.8 |
| PSA-GAN | 0.623(10) | 0.546(10) | 0.537(11) | 0.488(12) | 1.220(13) | 0.578(13) | 0.535(11) | 0.533(11) | 1.614(11) | 0.697(11) | 11.3 |
| Dlinear | 0.415(4) | 0.442(5) | 0.345(4) | 0.378(4) | 0.488(11) | 0.444(11) | 0.215(5) | 0.336(4) | **0.389**(1) | **0.268**(1) | 5.0 |
| PatchTST | 0.526(8) | 0.489(7) | 0.372(6) | 0.392(7) | 0.782(12) | 0.555(12) | 0.225(6) | 0.348(6) | 0.831(9) | 0.411(8) | 8.1 |
| Fedformer | 0.541(9) | 0.484(6) | 0.426(7) | 0.413(9) | 0.342(4) | 0.347(4) | 0.238(7) | 0.341(5) | 0.591(5) | 0.385(5) | 6.1 |
| Autoformer | 0.516(7) | 0.493(8) | 0.565(12) | 0.435(10) | 0.360(7) | 0.385(10) | 0.201(4) | 0.379(9) | 0.688(8) | 0.390(6) | 8.1 |
| Informer | 0.775(13) | 0.567(12) | 0.673(13) | 0.592(13) | 0.385(9) | 0.370(6) | 0.298(10) | 0.405(10) | 0.664(7) | 0.440(9) | 10.2 |

In addition, we compare CHIME with other time series models in multivariate time series forecasting tasks, as shown in Table 3. CHIME achieves the highest average ranking, surpassing both other generative models and diffusion-based time series models. Notably, CHIME demonstrates consistently lower MSE, particularly in complex scenarios such as Weather and Electricity datasets, indicating its higher prediction stability and improved accuracy in learning intricate patterns.

*4.6. CHIME: Superior Performance on Few-Shot Scenarios*

While diffusion models have demonstrated remarkable generative capabilities, they can still face limitations when the available data is insufficient. In the time series domain, few-shot learning scenarios are common, particularly in real-world applications. Limited data may only capture partial temporal features, and the observed distribution can shift over long time horizons or across different subjects, posing significant challenges for conventional diffusion paradigms.

Typically, for a new subject, a diffusion model would need to be retrained from scratch. However, if the available data for the new subject is scarce, directly applying a model trained on other similar subjects may result in distributional mismatch and poor downstream performance, whereas retraining may lead to the loss of long-term temporal features. Our proposed CHIME framework is well-suited for handling such few-shot scenarios. During pre-training, we equip the diffusion model with the ability to reconstruct multi-scale features based on conditional information. Subsequently, the feature hallucination module refines the limited conditional information.

The time series forecasting results under the few-shot setting are presented in Figure 6. We use 20% of the original dataset as the accessible training samples and set the forecasting horizon to $h = 96$. For other deep learning models, we first adopt the architectural parameters reported in their original papers and then perform one round of manual tuning to select the best-performing models. To ensure consistency, we implement a unified set of training hyperparameters across all datasets: all models are trained for 10 epochs with early stopping (patience = 5) based on validation loss, using the Adam optimizer with an initial learning rate of 0.0001. The batch size is set to 64.

It can be observed that CHIME achieves the lowest forecasting MSE across all five datasets, demonstrating outstanding robustness under few-shot conditions. In contrast, Diffusion-TS is more adversely affected by the reduced number of samples. Figure 7 presents a comprehensive evaluation plot comparing generative and predictive performance across different generative models, where CHIME consistently maintains strong generation and prediction capabilities under few-shot settings.

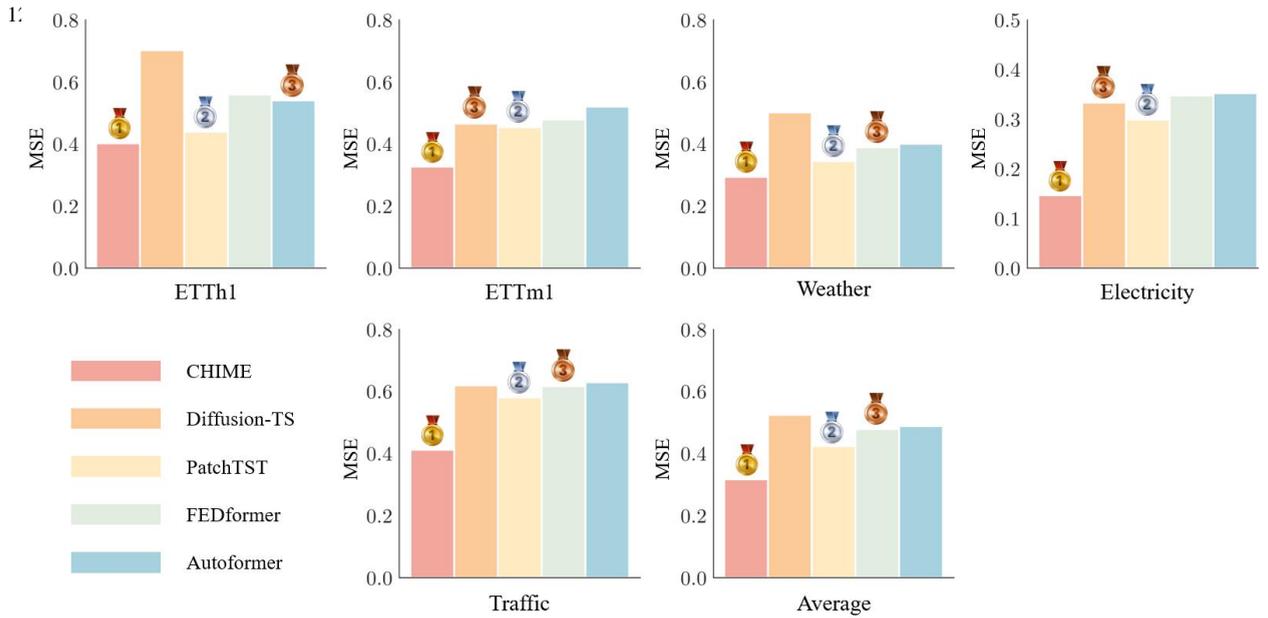

**Fig. 6** Time series forecasting results under few-shot setting

This strong few-shot adaptation ability is primarily attributed to the design of the FH module. Even when only limited data is available for a new subject, forming a few-shot setting, we assume that complete data from other similar subjects remains accessible. For example, if only a small amount of electricity load data is available for a new user, full data from other users (not necessarily from the same dataset) can still be utilized. The feature hallucination module, trained on these additional sources, enhances the alignment between the conditional information and the true distribution, thereby improving noise reconstruction accuracy during the denoising process.

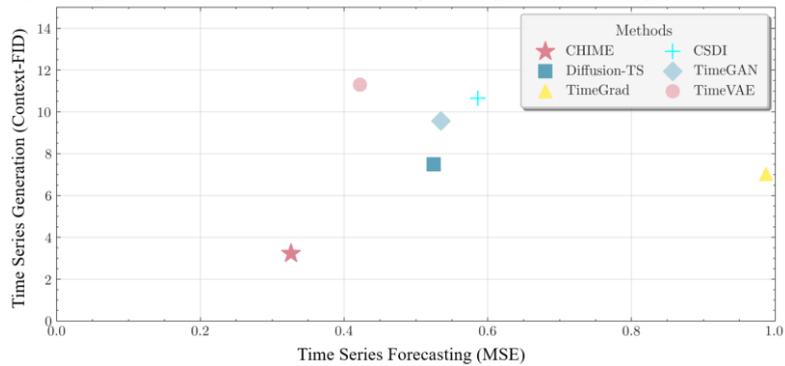

**Fig. 7** Comparison between CHIME and other generative baselines under few-shot setting

### 4.7. Ablation Study

**Model Architecture Ablation Study.** To validate the effectiveness of the components proposed in CHIME, we performed an ablation study by comparing the model's performance when removing the multi-scale integration module, using average weights, and removing the Feature Hallucination module. The test results for generation tasks, forecasting tasks, and both generation and forecasting tasks under the few-shot setting are shown in Table 4.

From the table, it is evident that the full CHIME architecture achieves optimal performance across all tasks. When the multi-scale integration module is removed, the model's error metrics significantly increase in both forecasting and generation tasks, indicating that this module effectively captures the characteristics of different time series components and ensures proper alignment of the in-domain distribution. When average weights are used, the model's performance decreases slightly, validating the rationale behind incorporating multi-scale information weighting. Removing the Feature Hallucination module results in a substantial increase in error under the few-shot scenario, underscoring the importance of conditional denoising through feature hallucination. Notably, in the experiment where the multi-scale integration module was removed, the model still outperforms Diffusion-TS in the few-shot setting, as the feature hallucination module helps reduce prediction errors (0.513 vs. 0.525).

**Table 4** Ablation study of CHIME architecture: Performance comparison of different configurations on generation and forecasting tasks, including few-shot learning settings.

|  | Gen. (Context-FID) | Pre. (MAE) | Gen. in FSL | Pre. in FSL (MAE) |
|---|---|---|---|---|
| CHIME | **0.075±.013** | **0.328** | **3.236±.009** | **0.413** |
| w/o multi-scale integration | 0.276±.048 | 0.447 | 5.996±.008 | 0.513 |
| average weight | 0.093±.020 | 0.349 | 3.742±.001 | 0.444 |
| w/o feature hallucination | 0.079±.014 | 0.350 | 8.208±.069 | 0.647 |

**Conditional Denoising Paradigm Ablation Study.** To validate the effectiveness of the conditional denoising training paradigm proposed in this paper, we conducted three comparative experiments: data reconstruction, attention with side information, and attention on original conditions. The results of each task are presented in Table 5.

**Table 5** Ablation study of conditional denoising paradigm

|  | Gen. (Context-FID) | Pre. (MAE) | Gen. in FSL | Pre. in FSL (MAE) |
|---|---|---|---|---|
| $\varepsilon$ -reconstruction (CHIME) | **0.075±.013** | **0.328** | **3.236±.009** | **0.413** |
| data-reconstruction | 0.085±.016 | 0.341 | 3.622±.009 | 0.425 |
| attention on $\varepsilon$ and condition (CHIME) | **0.075±.013** | **0.328** | **3.236±.009** | **0.413** |
| attention with side information (Liu et al., 2024) | 0.079±.012 | 0.334 | 3.787±.010 | 0.417 |
| attention on original condition | 0.079±.016 | 0.341 | 3.939±.007 | 0.434 |

It is evident that the proposed conditional denoising paradigm demonstrates best performance. Compared to focusing solely on the original conditional information or incorporating side information to attention mechanism, in our approach, the multi-head attention mechanism effectively guides the denoising process at each step by focusing on both $\varepsilon$ and the conditional information, thereby achieving more accurate generation.

**Feature Hallucination from a Semi-Supervised Perspective.** In Section 4.6, we discussed the advantages of CHIME in few-shot settings, which primarily stem from the feature hallucination-based conditional denoising paradigm. In the previous few-shot setup, we used 20% of the dataset. This prompted us to propose the following hypothesis: Can CHIME's few-shot learning capability be considered an effective model for semi-supervised tasks? To investigate this, we conducted a series of experiments with varying percentages of accessible data. The results are shown in Figure 8.

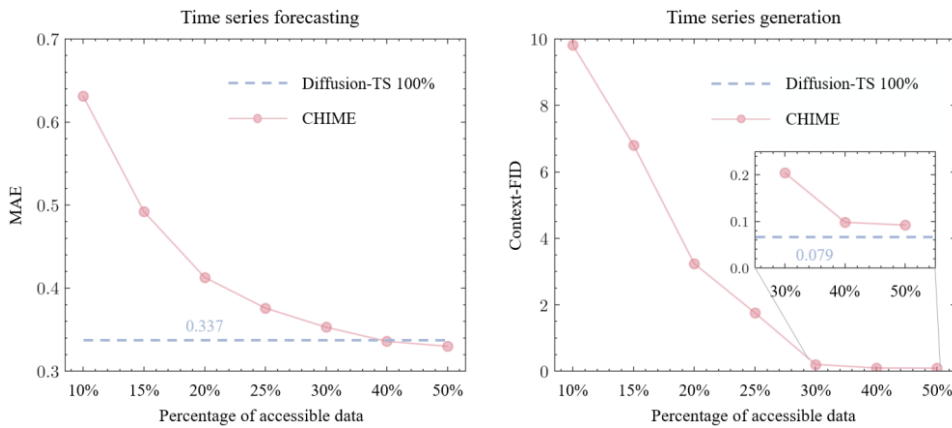

**Fig. 8** Comparison between CHIME and other generative baselines under few-shot setting

From Figure 8, it is clear that as the percentage of accessible data decreases, both the prediction errors and the generated Context-FID metrics for CHIME increase, with more drastic changes observed as the data size reduces. When the accessible data percentage reaches 40%, CHIME's performance on the forecasting task surpasses that of Diffusion-TS with 100% data, demonstrating its semi-supervised task capabilities. In the generation task, while CHIME does not outperform Diffusion-TS with 100% data when accessing only 50% of the data, it still shows a significant improvement compared to other baseline models under the same few-shot setting.

**Impact of Feature Hallucination Granularity.** In this paper, we propose Feature Hallucination within conditional denoising, which effectively enhances the alignment between the generated sequence feature distribution and the real sample distributions, thereby improving the model's generalization capability across subjects and over long-time scales. Clearly, selecting different granularities for Feature Hallucination can have varying effects on the final results. To investigate this, we designed a set of additional experiments to explore the impact of Feature Hallucination granularity on both prediction and generation performance, as well as its behavior under the few-shot setting.

**Table 6** Impact of feature hallucination granularity

| Granularity of FH | 1 day | 2 days | 3 days | **week** | half-month | month | season |
|---|---|---|---|---|---|---|---|
| Gen. (Context-FID) | $0.075 \pm .012$ | $0.075 \pm .012$ | $0.075 \pm .012$ | $0.075 \pm .013$ | $0.077 \pm .013$ | $0.079 \pm .013$ | $0.084 \pm .054$ |
| Pre. (MAE) | 0.327 | 0.327 | 0.328 | 0.328 | 0.341 | 0.347 | 0.376 |
| Gen. in FSL | $3.107 \pm .010$ | $3.107 \pm .010$ | $3.111 \pm .009$ | $3.236 \pm .009$ | $5.176 \pm .010$ | $7.220 \pm .024$ | $8.099 \pm .120$ |
| Pre. In FSL | 0.412 | 0.412 | 0.412 | 0.413 | 0.458 | 0.517 | 0.601 |

Table 6 presents the model's performance metrics for different granularities, namely 1 day, 2 days, 3 days, week, half-month, month, and season. It can be observed that as the granularity increases, CHIME's performance in all tasks steadily declines. When the granularity is smaller than one week, the change is not significant, but when it exceeds one week, the model's prediction error and other metrics start to rise sharply. Therefore, the main experiments in this paper select a granularity of one week. We did not choose finer granularities, as they offer minimal improvement to the model and would require significantly more training time. Notably, when a season granularity is used, the model's performance in both forecasting and generation tasks is even worse than the performance when the Feature Hallucination module is removed (see Table 4, "w/o feature hallucination"), suggesting that excessively coarse granularity may have an adverse effect on the model.

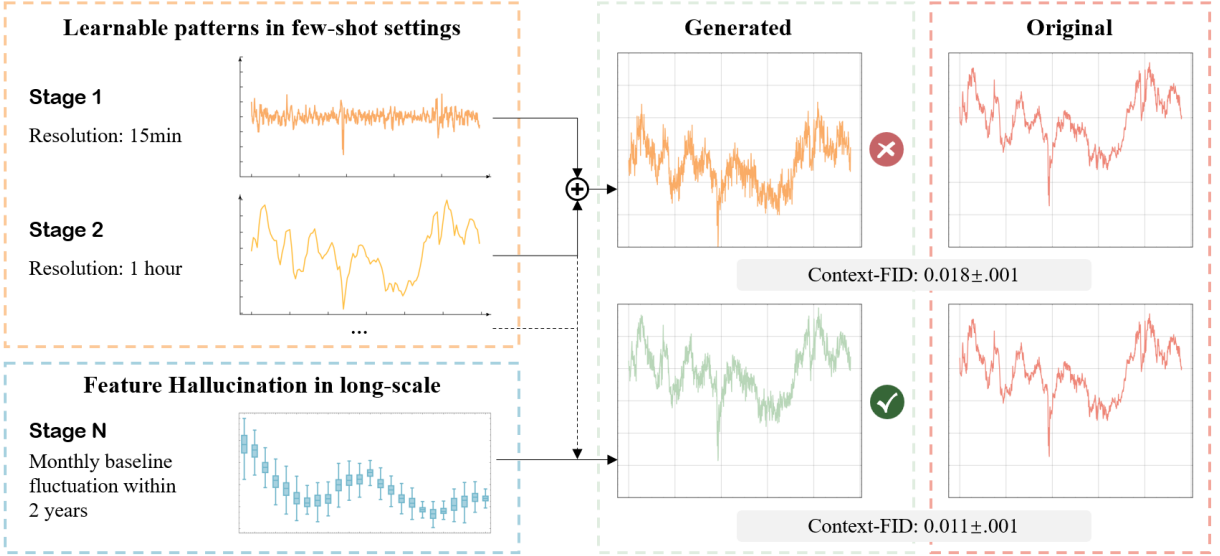

**Fig. 9** Case study on Ettm1

**Interpretability Analysis.** The model proposed in this paper exhibits good interpretability. Figure 9 presents a case study on the ETTM dataset, detailing the roles of our Feature Hallucination module and Multi-scale Integration framework in time series generation. In a few-shot scenario, only fine-grained sequences such as Stage 1, Stage 2, and so on can be generated, as shown in the left column of Figure 9. However, the trend variations of the time series over long time scales are neglected in the few-shot setup. As seen in the box plot for the monthly baseline level, the ETTM dataset shows an increasing trend from January to June, and a decreasing trend from July to December, with an overall downward trend. Without the Feature Hallucination module, the result after integrating fine-grained scale information, shown by the orange curve, replicates the local features of the original sequence but exhibits discrepancies at the overall baseline level. After adding the Feature Hallucination module, the conditionally transferred information is fed into the denoising network, leading to the generation of a more accurate sequence.

## 5. Conclusion

In this paper, we proposed CHIME, a **c**onditional **h**allucination and **i**ntegrated **m**ulti-scale **e**nhancement framework for time series diffusion models. To address the unique characteristics of time series data, we applied multi-scale decomposition and integration to obtain a multi-scale mixed representation, enabling the alignment of in-domain feature distributions between generated samples and real samples. We incorporated a Feature Hallucination module within the conditional denoising process to facilitate cross-subject and long-time scale feature transfer, thereby enhancing the model's generalization capability. Extensive experiments demonstrate that CHIME outperforms existing time series diffusion models.

## Acknowledgement

This work was supported by National Natural Science Foundation of China (U24B6010).

## References

Alaa, A., Chan, A. J., & van der Schaar, M. (2021). Generative time-series modeling with fourier flows. International Conference on Learning Representations,

Desai, A., Freeman, C., Wang, Z., & Beaver, I. TimeVAE: A Variational Auto-Encoder for Multivariate Time Series Generation.

Esteban, C., Hyland, S. L., & Rätsch, G. (2017). Real-valued (medical) time series generation with recurrent conditional gans. *arXiv preprint arXiv:1706.02633*.

Fan, X., Wu, Y., Xu, C., Huang, Y., Liu, W., & Bian, J. MG-TSD: Multi-Granularity Time Series Diffusion Models with Guided Learning Process. The Twelfth International Conference on Learning Representations,

Fang, W., Zhuo, W., Yan, J., Song, Y., Jiang, D., & Zhou, T. (2022). Attention meets long short-term memory: A deep learning network for traffic flow forecasting. *Physica A: Statistical Mechanics and its Applications*, *587*, 126485.

Guan, Y., Li, D., Xue, S., & Xi, Y. (2021). Feature-fusion-kernel-based Gaussian process model for probabilistic long-term load forecasting. *Neurocomputing*, *426*, 174-184. https://doi.org/https://doi.org/10.1016/j.neucom.2020.10.043

Hariharan, B., & Girshick, R. (2017). Low-shot visual recognition by shrinking and hallucinating features. Proceedings of the IEEE international conference on computer vision,

Ho, J., Jain, A., & Abbeel, P. (2020). Denoising diffusion probabilistic models. *Advances in Neural Information Processing Systems*, *33*, 6840-6851.

Hou, M., Xu, C., Liu, Y., Liu, W., Bian, J., Wu, L., Li, Z., Chen, E., & Liu, T.-Y. (2021). Stock trend prediction with multi-granularity data: A contrastive learning approach with adaptive fusion. Proceedings of the 30th ACM International Conference on Information & Knowledge Management,

Jeha, P., Bohlke-Schneider, M., Mercado, P., Kapoor, S., Nirwan, R. S., Flunkert, V., Gasthaus, J., & Januschowski, T. (2022). PSA-GAN: Progressive self attention GANs for synthetic time series. The Tenth International Conference on Learning Representations,

Liao, S., Ni, H., Szpruch, L., Wiese, M., Sabate-Vidales, M., & Xiao, B. (2020). Conditional sig-wasserstein gans for time series generation. *arXiv preprint arXiv:2006.05421*.

Lim, B., & Zohren, S. (2021). Time-series forecasting with deep learning: a survey. *Philosophical Transactions of the Royal Society A*, *379*(2194), 20200209.

Liu, J., Yang, L., Li, H., & Hong, S. (2024). Retrieval-augmented diffusion models for time series forecasting. *Advances in Neural Information Processing Systems*, *37*, 2766-2786.

Lopez Alcaraz, J. M., & Strodthoff, N. (2023). Diffusion-based time series imputation and forecasting with structured atate apace models. *Transactions on machine learning research*, 1-36.

Mogren, O. (2016). C-RNN-GAN: Continuous recurrent neural networks with adversarial training. *arXiv preprint arXiv:1611.09904*.

Nie, Y., Nguyen, N. H., Sinthong, P., & Kalagnanam, J. A Time Series is Worth 64 Words: Long-term Forecasting with Transformers. The Eleventh International Conference on Learning Representations,

Rajpurkar, P., Chen, E., Banerjee, O., & Topol, E. J. (2022). AI in health and medicine. *Nature medicine*, *28*(1), 31-38.

Rasul, K., Seward, C., Schuster, I., & Vollgraf, R. (2021). Autoregressive denoising diffusion models for multivariate probabilistic time series forecasting. International conference on machine learning,

Shen, L., Chen, W., & Kwok, J. (2024). Multi-resolution diffusion models for time series forecasting. The Twelfth International Conference on Learning Representations,

Shen, L., & Kwok, J. (2023). Non-autoregressive conditional diffusion models for time series prediction. International Conference on Machine Learning,

Tashiro, Y., Song, J., Song, Y., & Ermon, S. (2021). Csdi: Conditional score-based diffusion models for probabilistic time series imputation. *Advances in Neural Information Processing Systems*, *34*, 24804-24816.

Vaswani, A. (2017). Attention is all you need. *Advances in Neural Information Processing Systems*.

Wu, H., Xu, J., Wang, J., & Long, M. (2021). Autoformer: Decomposition transformers with auto-correlation for long-term series forecasting. *Advances in Neural Information Processing Systems*, *34*, 22419-22430.

Wu, H., Zhou, H., Long, M., & Wang, J. (2023). Interpretable weather forecasting for worldwide stations with a unified deep model. *Nature Machine Intelligence*, *5*(6), 602-611.

Yoon, J., Jarrett, D., & Van der Schaar, M. (2019). Time-series generative adversarial networks. *Advances in Neural Information Processing Systems*, *32*.

Yuan, X., & Qiao, Y. Diffusion-TS: Interpretable Diffusion for General Time Series Generation. The Twelfth International Conference on Learning Representations,

Zeng, A., Chen, M., Zhang, L., & Xu, Q. (2023). Are transformers effective for time series forecasting? Proceedings of the AAAI Conference on Artificial Intelligence,

Zhang, M., Guo, X., Pan, L., Cai, Z., Hong, F., Li, H., Yang, L., & Liu, Z. (2023). Remodiffuse: Retrieval-augmented motion diffusion model. Proceedings of the IEEE/CVF International Conference on Computer Vision,

Zhou, H., Zhang, S., Peng, J., Zhang, S., Li, J., Xiong, H., & Zhang, W. (2021). Informer: Beyond Efficient Transformer for Long Sequence Time-Series Forecasting. *Proceedings of the AAAI Conference on Artificial Intelligence*, *35*(12), 11106-11115. https://doi.org/10.1609/aaai.v35i12.17325

Zhou, T., Ma, Z., Wen, Q., Wang, X., Sun, L., & Jin, R. (2022). *FEDformer: Frequency Enhanced Decomposed Transformer for Long-term Series Forecasting* Proceedings of the 39th International Conference on Machine Learning, Proceedings of Machine Learning Research. https://proceedings.mlr.press/v162/zhou22g.html